\documentclass[11pt]{article}
\usepackage{eacl2014}
\usepackage{multirow}
\usepackage{multicol}
\usepackage{url}
\usepackage{latexsym}
\usepackage{times}
\usepackage{graphicx}
\usepackage{CJKutf8}
\usepackage{bm}

\title{Neural Reranking Improves Subjective Quality of Machine Translation: \\ NAIST at WAT2015}

\author{Graham Neubig, Makoto Morishita, Satoshi Nakamura \\
  Graduate School of Information Science \\
  Nara Institute of Science and Technology \\
  8916-5 Takayama-cho, Ikoma-shi, Nara, Japan \\
  {\tt \{neubig,morishita.makoto.mb1,s-nakamura\}@is.naist.jp}}

\date{}

\begin{document}
\maketitle
\begin{abstract}
This year, the Nara Institute of Science and Technology (NAIST)'s submission to the 2015 Workshop on Asian Translation was based on syntax-based statistical machine translation, with the addition of a reranking component using neural attentional machine translation models.
Experiments re-confirmed results from previous work stating that neural MT reranking provides a large gain in objective evaluation measures such as BLEU, and also confirmed for the first time that these results also carry over to manual evaluation.
We further perform a detailed analysis of reasons for this increase, finding that the main contributions of the neural models lie in improvement of the grammatical correctness of the output, as opposed to improvements in lexical choice of content words.
\end{abstract}

\section{Introduction}
\label{sec:intro}

Neural network models for machine translation (MT) \cite{kalchbrenner13rnntm,sutskever14sequencetosequence,bahdanau15alignandtranslate}, while still in a nascent stage, have shown impressive results in a number of translation tasks.
Specifically, a number of works have demonstrated gains in BLEU score \cite{papineni02bleu} over state-of-the-art non-neural systems, both when using the neural MT model stand-alone \cite{luong15rareword,jean15verylarge,luong15effectiveattentional}, or to rerank the output of more traditional systems phrase-based MT systems \cite{sutskever14sequencetosequence}.

However, despite these impressive results with regards to automatic measures of translation quality, there has been little examination of the effect that these gains have on the subjective impressions of human users.
Because BLEU generally has some correlation with translation quality,\footnote{Particularly when comparing similar systems, such as the case of when neural MT is used for reranking existing system results.} it is fair to hypothesize that these gains will carry over to gains in human evaluation, but empirical evidence for this hypothesis is still scarce.
In this paper, we attempt to close this gap by examining the gains provided by using neural MT models to rerank the hypotheses a state-of-the-art non-neural MT system, both from the objective and subjective perspectives.

Specifically, as part of the Nara Institute of Science and Technology (NAIST) submission to the Workshop on Asian Translation (WAT) 2015 \cite{nakazawa15wat}, we generate reranked and non-reranked translation results in four language pairs (Section \ref{sec:generation}).
Based on these translation results, we calculate scores according to automatic evaluation measures BLEU and RIBES \cite{isozaki10ribes}, and a manual evaluation that involves comparing hypotheses to a baseline system (Section \ref{sec:evaluation}).
Next, we perform a detailed analysis of the cases in which subjective impressions improved or degraded due to neural MT reranking, and identify major areas in which neural reranking improves results, and areas in which reranking is less helpful (Section \ref{sec:analysis}).
Finally, as an auxiliary result, we also examine the effect that the size of the $n$-best list used in reranking has on the improvement of translation results (Section \ref{sec:nbestsize}).

\section{Generation of Translation Results}
\label{sec:generation}

\subsection{Baseline System}
\label{sec:baseline}

All experiments are performed on WAT2015 translation task from Japanese (ja) to/from English (en) and Chinese (zh).
As a baseline, we used the NAIST system for WAT 2014 \cite{neubig14wat}, a state-of-the-art system that achieved the highest accuracy on all four tracks in the last year's evaluation.\footnote{Scripts to reproduce the system are available at \url{http://phontron.com/project/wat2014}.}
The details of construction are described in \newcite{neubig14wat}, but we briefly outline it here for completeness.

The system is based on the Travatar toolkit \cite{neubig13travatar}, using tree-to-string statistical MT \cite{graehl04treetransducers,liu06treetostring}, in which the source is first syntactically parsed, then subtrees of the input parse are converted into strings on the target side.
This translation paradigm has proven effective for translation between syntactically distant language pairs such as those handled by the WAT tasks.
In addition, following our findings in \newcite{neubig14acl}, to improve the accuracy of translation we use forest-based encoding of many parse candidates \cite{mi08forestbased}, and a supervised alignment technique for ja-en and en-ja \cite{riesa10hierarchicalwordalignment}.

To train the systems, we used the ASPEC corpus provided by WAT.
For the zh-ja and ja-zh systems, we used all of the data, amounting to 672k sentences.
For the en-ja and ja-en systems, we used all 3M sentences for training the language models, and the first 2M sentences of the training data for training the translation models.

For English, Japanese, and Chinese, tokenization was performed using the Stanford Parser \cite{klein03accurateunlexicalized}, the KyTea toolkit \cite{neubig11aclshort}, and the Stanford Segmenter \cite{tseng05crfws} respectively.
For parsing, we use the Egret parser,\footnote{\url{https://github.com/neubig/egret}} which implements the latent variable parsing model of \cite{petrov06lapcfg}.%
\footnote{In addition, for ja-en translation, we make one modification to the parser used in the previous year's submission, performing parser self-training \cite{mcclosky06effectiveselftraining} using sentences from the training data that had a BLEU score greater than 0.8, and selecting the tree corresponding to the 500-best hypothesis that had the best score according to BLEU+1 \cite{lin04orange}.}

For all systems, we trained a 6-gram language model smoothed with modified Kneser-Ney smoothing \cite{chen96smoothing} using KenLM \cite{heafield13estimation}.
To optimize the parameters of the log-linear model, we use standard minimum error rate training (MERT; \newcite{och03mert}) with BLEU as an objective.

\subsection{Neural MT Models}
\label{sec:neuralmt}

As our neural MT model, we use the attentional model of \newcite{bahdanau15alignandtranslate}.
The model first encodes the source sentence $\bm{f}$ using bidirectional long short-term memory (LSTM; \newcite{hochreiter97lstm}) recurrent networks.
This results in an encoding vector $\bm{h}_j$ for each word $f_j$ in $\bm{f}$.
The model then proceeds to generate the target translation $\hat{\bm{e}}$ one word at a time, at each time step calculating soft alignments $\bm{a}_i$ that are used to generate a context vector $\bm{g}_i$, which is referenced when generating the target word
\begin{equation}
\bm{g}_i = \sum_{j=1}^{|\bm{f}|} a_{i,j} \bm{h_j}.
\end{equation}

Attentional models have a number of appealing properties, such as being theoretically able to encode variable length sequences without worrying about memory constraints imposed by the fixed-size vectors used in encoder-decoder models.
These advantages are confirmed in empirical results, with attentional models performing markedly better on longer sequences \cite{bahdanau15alignandtranslate}.

To train the neural MT models, we used the implementation provided by the lamtram toolkit.\footnote{\url{http://github.com/neubig/lamtram}}
The forward and reverse LSTM models each had 256 nodes, and word embeddings were also set to size 256.
For ja-en and en-ja models we chose the first 500k sentences in the training corpus, and for ja-zh and zh-ja models we used all 672k sentences. 
Training was performed using stochastic gradient descent (SGD) with an initial learning rate of 0.1, which was halved every epoch in which the development likelihood decreased.

For each language pair, we trained two models and ensembled the probabilities by linearly interpolating between the two probability distributions.\footnote{More standard log-linear interpolation resulted in similar, or slightly inferior results.}
These probabilities were used to rerank unique 1,000-best lists from the baseline model.
To perform reranking, the log likelihood of the neural MT model was added as an additional feature to the standard baseline model features, and the weight of this feature was decided by running MERT on the dev set.

\section{Experimental Results}
\label{sec:evaluation}

\begin{table*}[t]
\begin{center}
\begin{tabular}{l||rrr|rrr|rrr|rrr}
         & \multicolumn{3}{c|}{en-ja} & \multicolumn{3}{c|}{ja-en} & \multicolumn{3}{c|}{zh-ja} & \multicolumn{3}{c}{ja-zh} \\
System   & B      & R      & H        & B      & R      & H        & B      & R      & H        & B      & R      & H        \\ \hline \hline
Base     & 36.6   & 79.6   & 49.8     & 22.6   & 72.3   & 11.8     & 40.5   & 83.4   & 25.8     & 30.1   & 81.5   & 2.8      \\
Rerank   & \textbf{38.2}   & \textbf{81.4}   & \textbf{62.3}     & \textbf{25.4}   & \textbf{75.0}   & \textbf{35.5}     & \textbf{43.0}   & \textbf{84.8}   & \textbf{35.8}     & \textbf{31.6}   & \textbf{83.3}   & 7.0      \\
\end{tabular}
\end{center}
\caption{\label{tab:overall} Overall BLEU, RIBES, and HUMAN scores for our baseline system and system with neural MT reranking. Bold indicates a significant improvement according to bootstrap resampling at $p<0.05$ \cite{koehn04sigtest}.}
\end{table*}

First, we calculate overall numerical results for our systems with and without the neural MT reranking model.
As automatic evaluation we use the standard BLEU \cite{papineni02bleu} and reordering-oriented RIBES \cite{isozaki10ribes} metrics.
In manual evaluation, we use the WAT ``HUMAN'' evaluation score \cite{nakazawa15wat}, which is essentially related to the number of wins over a baseline phrase-based system.
In the case that the system beats the baseline on all sentences, the HUMAN score will be 100, and if it loses on all sentences the score will be -100.

From the results in Table \ref{tab:overall}, we can first see that adding the neural MT reranking resulted in a significant increase in the evaluation scores for all language pairs under consideration, except for the manual evaluation in ja-zh translation.%
\footnote{The overall scores for ja-zh are lower than others, perhaps a result of word-order between Japanese and Chinese being more similar than Japanese and English, the parser for Japanese being weaker than that of the other languages, and less consistent evaluation scores for the Chinese output \cite{nakazawa14wat}.}
It should be noted that these gains are achieved even though the original baseline was already quite strong (outperforming most other WAT2015 systems without a neural component).
While neural MT reranking has been noted to improve traditional systems with respect to BLEU score in previous work \cite{sutskever14sequencetosequence}, to our knowledge this is the first work that notes that these gains also carry over convincingly to human evaluation scores.
In the following section, we will examine the results in more detail and attempt to explain exactly what is causing this increase in translation quality.

\section{Analysis}
\label{sec:analysis}

\begin{table}[t]
\begin{center}
\begin{tabular}{l||rr|r}
\multicolumn{1}{c||}{Type} & \multicolumn{1}{c}{Impr.} & \multicolumn{1}{c|}{Degr.} & \multicolumn{1}{c}{\% Impr.} \\ \hline \hline
Reordering  & 55 & 9  & 86\% \\
Deletion    & 20 & 10 & 67\% \\
Insertion   & 19 & 2  & 90\% \\
Substitution& 15 & 11 & 58\% \\
Conjugation & 8  & 1  & 89\% \\ \hline
Total       & 117& 33 & 78\% \\
\end{tabular}
\end{center}
\caption{\label{tab:rough} A summary of the improvements and degradations caused by neural reranking.}
\end{table}

To perform a deeper analysis, we manually examined the first 200 sentences of the ja-en part of the official WAT2015 human evaluation set.
Specifically, we (1) compared the baseline and reranked outputs, and decided whether one was better or if they were of the same quality and (2) in the case that one of the two was better, classified the example by the type of error that was fixed or caused by the reranking leading to this change in subjective impression.
Specifically, when annotating the type of error, we used a simplified version of the error typology of \newcite{vilar06erroranalysis} consisting of \textit{insertion}, \textit{deletion}, \textit{word conjugation}, \textit{word substitution}, and \textit{reordering}, as well as subcategories of each of these categories (the number of sub-categories totalled approximately 40).
If there was more than one change in the sentence, only the change that we subjectively felt had the largest effect on the translation quality was annotated.

The number of improvements and degradations afforded by neural MT reranking is shown in Table \ref{tab:rough}.
From this figure, we can see that overall, neural reranking caused an improvement in 117 sentences, and a degradation in 33 sentences, corroborating the fact that the reranking process is giving consistent improvements in accuracy.
Further breaking down the changes, we can see that improvements in word reordering are by far the most prominent, slightly less than three times the number of improvements in the next most common category.
This demonstrates that the neural MT model is successfully capturing the overall structure of the sentence, and effectively disambiguating reorderings that could not be appropriately scored in the baseline model.

\begin{table*}[t]
\begin{center}
\small{
\begin{tabular}{|p{0.8cm}p{14cm}|} \hline
\multicolumn{2}{|l|}{\underline{1. Reordering of Phrases (+26, -4)}} \\
In.    & \begin{CJK}{UTF8}{min}症例２においては，直腸がんの肝転移に対する化学療法中に，発赤，硬結，皮膚潰ようを生じた。\end{CJK} \\
Ref.   & In case 2, reddening, induration, and skin ulcer appeared during chemical therapy for liver metastasis of rectal cancer. \\
Base.  & In case 2, occurred during chemotherapy for liver metastasis of rectal cancer, flare, induration, skin ulcer. \\
Rerank & In case 2, the flare, induration, skin ulcer was produced during the chemotherapy for hepatic metastasis of rectal cancer. \\ \hline
\multicolumn{2}{|l|}{\underline{2. Insertion/Deletion of Auxiliary Verbs (+15, -0)}} \\
In.    & \begin{CJK}{UTF8}{min}これにより得られる支配方程式は壁面乱流のようなせん断乱流にも有用である。\end{CJK} \\
Ref.   & Governing equation derived by this method is useful for turbulent shear flow like turbulent flow near wall. \\
Base.  & The governing equation \textbf{is} obtained by this is also useful for such as wall turbulence shear flow. \\
Rerank & The governing equation obtained by this is also useful for shear flow such as wall turbulence. \\ \hline
\multicolumn{2}{|l|}{\underline{3. Reordering of Coordinate Structures (+13, -2)}} \\
In.    & \begin{CJK}{UTF8}{min}レーザー加工は高密度光束による局所的な加熱とアブレーションにより行う。\end{CJK} \\
Ref.   & Laser work is done by local heating and ablation with high density light flux. \\
Base.  & The laser processing is carried out by local heating by high-density luminous flux and ablation. \\
Rerank & The laser processing is carried out by local heating and ablation by high-density flux. \\ \hline
\multicolumn{2}{|l|}{\underline{4. Conjugation of Verb Agreement (+6, -0)}} \\
In.    & \begin{CJK}{UTF8}{min}ラングミュア‐ブロジェット法や包接化にも触れた。\end{CJK} \\
Ref.   & Langmuir-Blodgett method and inclusion compounds are mentioned. \\
Base.  & Langmuir-Blodgett method and inclusion is also discussed. \\
Rerank & Langmuir-Blodgett method and inclusion are also mentioned. \\ \hline
\end{tabular}
}
\end{center}
\caption{\label{tab:examp} An example of more common varieties of improvements caused by the neural MT reranking.}
\end{table*}

Next in Table \ref{tab:examp} we show examples of the four most common sub-categories of errors that were fixed by the neural MT reranker, and note the total number of improvements and degradations of each.
The first subcategory is related to the general reordering of phrases in the sentence.
As there is a large amount of reordering involved in translating from Japanese to English, mistaken long-distance reordering is one of the more common causes for errors, and the neural MT model was effective at fixing these problems, resulting in 26 improvements and only 4 degradations.
In the sentence shown in the example, the baseline system swaps the verb phrase and subject positions, making it difficult to tell that the list of conditions are what ``occurred,'' while the reranked system appropriately puts this list as the subject of ``occurred.''

The second subcategory includes insertions or deletions of auxiliary verbs, for which there were 15 improvements and not a single degradation.
The reason why these errors occurred in the first place is that when a transitive verb, for example ``obtained,'' occurs on its own, it is often translated as ``X was obtained by Y,''\footnote{This passivization is somewhat of a trait of the scientific paper material used as material for this analysis.} but when it occurs as a relative clause decorating the noun X it will be translated as ``X obtained by Y,'' as shown in the example.
The baseline system does not include any explicit features to make this distinction between whether a verb is part of a relative clause or not, and thus made a number of mistakes of this variety.
However, it is evident that the neural MT model has learned to make this distinction, greatly reducing the number of these errors.

\begin{figure*}[t!]
\begin{center}
\includegraphics[width=160mm]{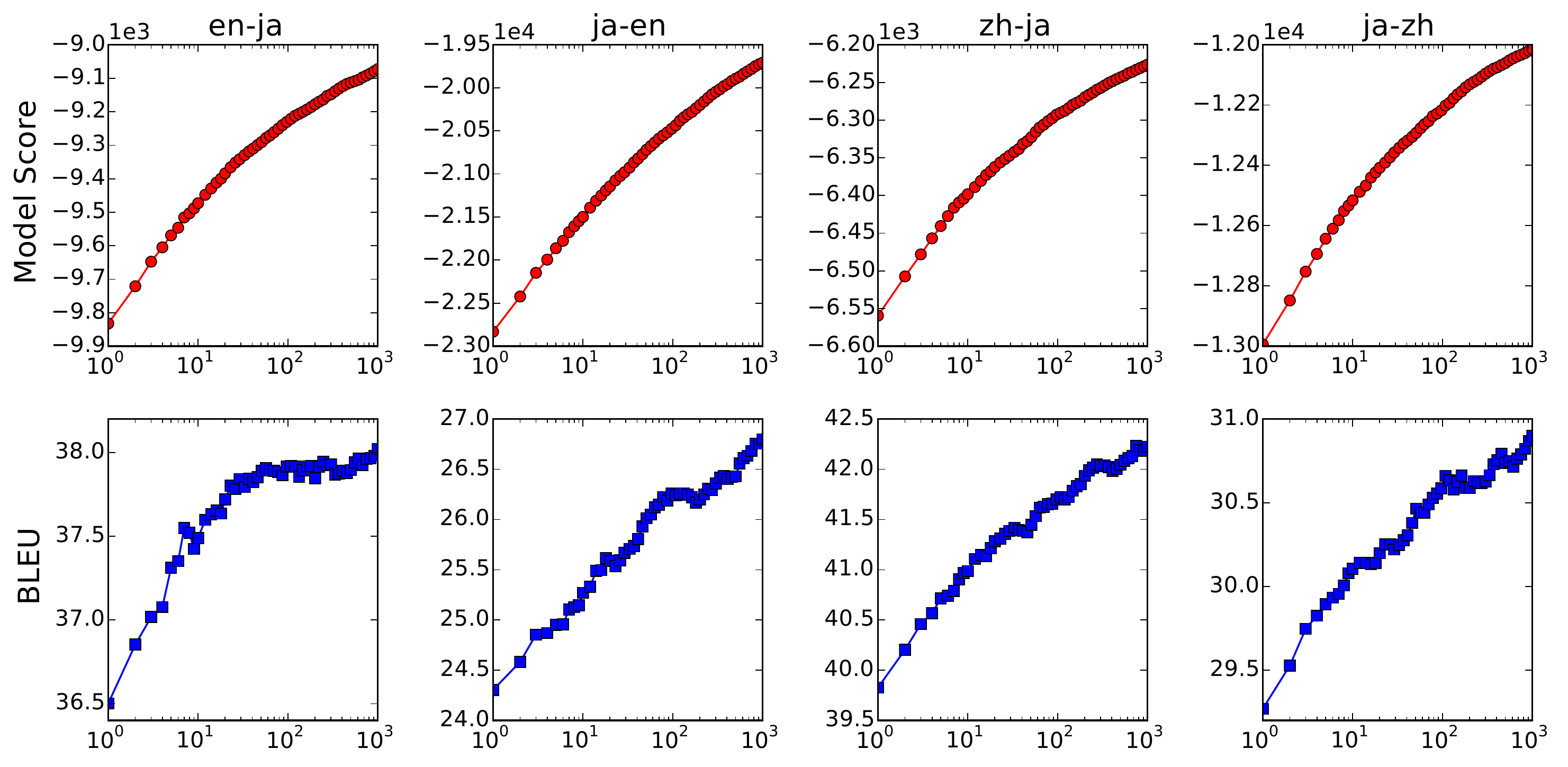}
\end{center}
\caption{Model and BLEU scores after neural MT reranking for each $n$-best list size (log scale).}
\label{fig:allscore}
\end{figure*}

The third subcategory is similar to the first, but explicitly involves the correct interpretation of coordinate structures.
It is well known that syntactic parsers often make mistakes in their interpretation of coordinate structures \cite{kummerfeld12wallstreetcorral}.
Of course, the parser used in our syntax-based MT system is no exception to this rule, and parse errors often cause coordinate phrases to be broken apart on the target side, as is the case in the example's ``local heating and ablation.''
The fact that the neural MT models were able to correct a large number of errors related to these structures suggests that they are able to successfully determine whether two phrases are coordinated or not, and keep them together on the target side.

The final sub-category of the top four is related to verb conjugation agreement.
Many of the examples related to verb conjugation, including the one shown in Table \ref{tab:examp}, were related to when two singular nouns were connected by a conjunction.
In this case, the local context provided by a standard $n$-gram language model is not enough to resolve the ambiguity, but the longer context handled by the neural MT model is able to resolve this easily.

What is notable about these four categories is that they all are related to improving the correctness of the output from a grammatical point of view, as opposed to fixing mistakes in lexical choice or terminology.
In fact, neural MT reranking had an overall negative effect on choice of terminology with only 2 improvements at the cost of 4 degradations.
This was due to the fact that the neural MT model tended to prefer more common words, mistaking ``radiant heat'' as ``radiation heat'' or ``slipring'' as ``ring.''
While these tendencies will be affected by many factors such as the size of the vocabulary or the number and size of hidden layers of the net, we feel it is safe to say that neural MT reranking can be expected to have a large positive effect on syntactic correctness of output, while results for lexical choice are less conclusive.

\section{Effect of $n$-best Size on Reranking}
\label{sec:nbestsize}

In the previous sections, we confirmed the effectiveness of $n$-best list reranking using neural MT models.
However, reranking using $n$-best lists (like other search methods for MT) is an approximate search method, and its effectiveness is limited by the size of the $n$-best list used.
In order to quantify the effect of this inexact search, we performed experiments to examine the post-reranking automatic evaluation scores of the MT results for all $n$-best list sizes from 1 to 1000.
Figure \ref{fig:allscore} shows the results of this examination, with the x-axis referring to the log-scaled number of hypotheses in the $n$-best list, and the y-axis referring to the quality of the translation, either with regards to model score (for the model including the neural MT likelihood as a feature) or BLEU score.\footnote{The BLEU scores differ slightly from Table \ref{tab:overall} due to differences in tokenization standards between these experiments and the official evaluation server.}

From these results we can note several interesting points.
First, we can see that the improvement in scores is very slightly sub-linear in the log number of hypotheses in the $n$-best list.
In other words, every time we double the $n$-best list size we will see an improvement in accuracy that is slightly smaller than the last time we doubled the size.
Second, we can note that in most cases this trend continues all the way up to our limit of 1000-best lists, indicating that gains are not saturating, and we can likely expect even more improvements from using larger lists, or perhaps directly performing decoding using neural models \cite{alkhouli15rnndecoding}.
The en-ja results, however, are an exception to this rule, with BLEU gains more or less saturating around the 50-best list point.

\section{Conclusion}
\label{sec:conclusion}

In this paper we described results applying neural MT reranking to a baseline syntax-based machine translation system in 4 languages.
In particular, we performed an in-depth analysis of what kinds of translation errors were fixed by neural MT reranking.
Based on this analysis, we found that the majority of the gains were related to improvements in the accuracy of transfer of correct grammatical structure to the target sentence, with the most prominent gains being related to errors regarding reordering of phrases, insertion/deletion of copulas, coordinate structures, and verb agreement.
We also found that, within the neural MT reranking framework, accuracy gains scaled approximately log-linearly with the size of the $n$-best list, and in most cases were not saturated even after examining 1000 unique hypotheses.

\section*{Acknowledgments:} This work was supported by JSPS KAKENHI Grant Number 25730136.

\bibliographystyle{acl}
\bibliography{myabbrv,main}

\begin{thebibliography}{}

\bibitem[\protect\citename{Alkhouli \bgroup et al.\egroup
  }2015]{alkhouli15rnndecoding}
Tamer Alkhouli, Felix Rietig, and Hermann Ney.
\newblock 2015.
\newblock Investigations on phrase-based decoding with recurrent neural network
  language and translation models.
\newblock In {\em Proc. WMT}, pages 294--303.

\bibitem[\protect\citename{Bahdanau \bgroup et al.\egroup
  }2015]{bahdanau15alignandtranslate}
Dzmitry Bahdanau, Kyunghyun Cho, and Yoshua Bengio.
\newblock 2015.
\newblock Neural machine translation by jointly learning to align and
  translate.
\newblock In {\em Proc. ICLR}.

\bibitem[\protect\citename{Chen and Goodman}1996]{chen96smoothing}
Stanley~F. Chen and Joshua Goodman.
\newblock 1996.
\newblock An empirical study of smoothing techniques for language modeling.
\newblock In {\em Proc. ACL}, pages 310--318.

\bibitem[\protect\citename{Graehl and Knight}2004]{graehl04treetransducers}
Jonathan Graehl and Kevin Knight.
\newblock 2004.
\newblock Training tree transducers.
\newblock In {\em Proc. HLT}, pages 105--112.

\bibitem[\protect\citename{Heafield \bgroup et al.\egroup
  }2013]{heafield13estimation}
Kenneth Heafield, Ivan Pouzyrevsky, Jonathan~H. Clark, and Philipp Koehn.
\newblock 2013.
\newblock Scalable modified {Kneser-Ney} language model estimation.
\newblock In {\em Proc. ACL}, pages 690--696.

\bibitem[\protect\citename{Hochreiter and Schmidhuber}1997]{hochreiter97lstm}
Sepp Hochreiter and J{\"u}rgen Schmidhuber.
\newblock 1997.
\newblock Long short-term memory.
\newblock {\em Neural computation}, 9(8):1735--1780.

\bibitem[\protect\citename{Isozaki \bgroup et al.\egroup }2010]{isozaki10ribes}
Hideki Isozaki, Tsutomu Hirao, Kevin Duh, Katsuhito Sudoh, and Hajime Tsukada.
\newblock 2010.
\newblock Automatic evaluation of translation quality for distant language
  pairs.
\newblock In {\em Proc. EMNLP}, pages 944--952.

\bibitem[\protect\citename{Jean \bgroup et al.\egroup }2015]{jean15verylarge}
S{\'{e}}bastien Jean, Kyunghyun Cho, Roland Memisevic, and Yoshua Bengio.
\newblock 2015.
\newblock On using very large target vocabulary for neural machine translation.
\newblock In {\em Proc. ACL}, pages 1--10.

\bibitem[\protect\citename{Kalchbrenner and Blunsom}2013]{kalchbrenner13rnntm}
Nal Kalchbrenner and Phil Blunsom.
\newblock 2013.
\newblock Recurrent continuous translation models.
\newblock In {\em Proc. EMNLP}, pages 1700--1709.

\bibitem[\protect\citename{Klein and
  Manning}2003]{klein03accurateunlexicalized}
Dan Klein and Christopher~D. Manning.
\newblock 2003.
\newblock Accurate unlexicalized parsing.
\newblock In {\em Proc. ACL}, pages 423--430.

\bibitem[\protect\citename{Koehn}2004]{koehn04sigtest}
Philipp Koehn.
\newblock 2004.
\newblock Statistical significance tests for machine translation evaluation.
\newblock In {\em Proc. EMNLP}, pages 388--395.

\bibitem[\protect\citename{Kummerfeld \bgroup et al.\egroup
  }2012]{kummerfeld12wallstreetcorral}
Jonathan~K Kummerfeld, David Hall, James~R Curran, and Dan Klein.
\newblock 2012.
\newblock Parser showdown at the wall street corral: an empirical investigation
  of error types in parser output.
\newblock In {\em Proc. EMNLP}, pages 1048--1059.

\bibitem[\protect\citename{Lin and Och}2004]{lin04orange}
Chin-Yew Lin and Franz~Josef Och.
\newblock 2004.
\newblock Orange: a method for evaluating automatic evaluation metrics for
  machine translation.
\newblock In {\em Proc. COLING}, pages 501--507.

\bibitem[\protect\citename{Liu \bgroup et al.\egroup }2006]{liu06treetostring}
Yang Liu, Qun Liu, and Shouxun Lin.
\newblock 2006.
\newblock Tree-to-string alignment template for statistical machine
  translation.
\newblock In {\em Proc. ACL}, pages 609--616.

\bibitem[\protect\citename{Luong \bgroup et al.\egroup }2015a]{luong15rareword}
Minh-Thang Luong, Ilya Sutskever, Quoc Le, Oriol Vinyals, and Wojciech Zaremba.
\newblock 2015a.
\newblock Addressing the rare word problem in neural machine translation.
\newblock In {\em Proc. ACL}, pages 11--19.

\bibitem[\protect\citename{Luong \bgroup et al.\egroup
  }2015b]{luong15effectiveattentional}
Thang Luong, Hieu Pham, and Christopher~D. Manning.
\newblock 2015b.
\newblock Effective approaches to attention-based neural machine translation.
\newblock In {\em Proc. EMNLP}, pages 1412--1421.

\bibitem[\protect\citename{McClosky \bgroup et al.\egroup
  }2006]{mcclosky06effectiveselftraining}
David McClosky, Eugene Charniak, and Mark Johnson.
\newblock 2006.
\newblock Effective self-training for parsing.
\newblock In {\em Proc. HLT}, pages 152--159.

\bibitem[\protect\citename{Mi \bgroup et al.\egroup }2008]{mi08forestbased}
Haitao Mi, Liang Huang, and Qun Liu.
\newblock 2008.
\newblock Forest-based translation.
\newblock In {\em Proc. ACL}, pages 192--199.

\bibitem[\protect\citename{Nakazawa \bgroup et al.\egroup }2014]{nakazawa14wat}
Toshiaki Nakazawa, Hideki Mino, Isao Goto, Sadao Kurohashi, and Eiichiro
  Sumita.
\newblock 2014.
\newblock Overview of the 1st {Workshop on Asian Translation}.
\newblock In {\em Proc. WAT}.

\bibitem[\protect\citename{Nakazawa \bgroup et al.\egroup }2015]{nakazawa15wat}
Toshiaki Nakazawa, Hideya Mino, Isao Goto, Graham Neubig, Sadao Kurohashi, and
  Eiichiro Sumita.
\newblock 2015.
\newblock Overview of the 2nd {Workshop on Asian Translation}.
\newblock In {\em Proc. WAT}.

\bibitem[\protect\citename{Neubig and Duh}2014]{neubig14acl}
Graham Neubig and Kevin Duh.
\newblock 2014.
\newblock On the elements of an accurate tree-to-string machine translation
  system.
\newblock In {\em Proc. ACL}, pages 143--149.

\bibitem[\protect\citename{Neubig \bgroup et al.\egroup
  }2011]{neubig11aclshort}
Graham Neubig, Yosuke Nakata, and Shinsuke Mori.
\newblock 2011.
\newblock Pointwise prediction for robust, adaptable {Japanese} morphological
  analysis.
\newblock In {\em Proc. ACL}, pages 529--533.

\bibitem[\protect\citename{Neubig}2013]{neubig13travatar}
Graham Neubig.
\newblock 2013.
\newblock Travatar: A forest-to-string machine translation engine based on tree
  transducers.
\newblock In {\em Proc. ACL Demo Track}, pages 91--96.

\bibitem[\protect\citename{Neubig}2014]{neubig14wat}
Graham Neubig.
\newblock 2014.
\newblock Forest-to-string {SMT} for {Asian} language translation: {NAIST} at
  {WAT2014}.
\newblock In {\em Proc. WAT}.

\bibitem[\protect\citename{Och}2003]{och03mert}
Franz~Josef Och.
\newblock 2003.
\newblock Minimum error rate training in statistical machine translation.
\newblock In {\em Proc. ACL}, pages 160--167.

\bibitem[\protect\citename{Papineni \bgroup et al.\egroup
  }2002]{papineni02bleu}
Kishore Papineni, Salim Roukos, Todd Ward, and Wei-Jing Zhu.
\newblock 2002.
\newblock {BLEU: a method for automatic evaluation of machine translation}.
\newblock In {\em Proc. ACL}, pages 311--318.

\bibitem[\protect\citename{Petrov \bgroup et al.\egroup }2006]{petrov06lapcfg}
Slav Petrov, Leon Barrett, Romain Thibaux, and Dan Klein.
\newblock 2006.
\newblock Learning accurate, compact, and interpretable tree annotation.
\newblock In {\em Proc. ACL}, pages 433--440.

\bibitem[\protect\citename{Riesa and
  Marcu}2010]{riesa10hierarchicalwordalignment}
Jason Riesa and Daniel Marcu.
\newblock 2010.
\newblock Hierarchical search for word alignment.
\newblock In {\em Proc. ACL}, pages 157--166.

\bibitem[\protect\citename{Sutskever \bgroup et al.\egroup
  }2014]{sutskever14sequencetosequence}
Ilya Sutskever, Oriol Vinyals, and Quoc~VV Le.
\newblock 2014.
\newblock Sequence to sequence learning with neural networks.
\newblock In {\em Proc. NIPS}, pages 3104--3112.

\bibitem[\protect\citename{Tseng \bgroup et al.\egroup }2005]{tseng05crfws}
Huihsin Tseng, Pichuan Chang, Galen Andrew, Daniel Jurafsky, and Christopher
  Manning.
\newblock 2005.
\newblock A conditional random field word segmenter for {SIGHAN} bakeoff 2005.
\newblock In {\em Proc. SIGHAN}.

\bibitem[\protect\citename{Vilar \bgroup et al.\egroup
  }2006]{vilar06erroranalysis}
David Vilar, Jia Xu, Luis~Fernando d’Haro, and Hermann Ney.
\newblock 2006.
\newblock Error analysis of statistical machine translation output.
\newblock In {\em Proc. LREC}, pages 697--702.

\end{thebibliography}

\end{document}